%% file: arxiv-submission.tex
\documentclass[final]{cvpr}

\usepackage{times}
\usepackage{epsfig}
\usepackage{graphicx}
\usepackage{amsmath}
\usepackage{amssymb}
\usepackage{bbold}

\usepackage{float}
\usepackage{dblfloatfix}

\usepackage[table]{xcolor}

\usepackage{booktabs}
\usepackage{multirow}

\usepackage{blindtext}

\makeatletter
\@namedef{ver@everyshi.sty}{}
\makeatother
\usepackage{pgfplots}
\pgfplotsset{compat=1.7}
\usepgfplotslibrary{groupplots}
\usetikzlibrary{positioning}
\usetikzlibrary{patterns}

\usepackage{url}

\usepackage[page,title]{appendix}

\usepackage[pagebackref=true,breaklinks=true,colorlinks,bookmarks=false]{hyperref}

\def\cvprPaperID{3} 
\def\confYear{CVPR 2021}
\begin{document}

\title{Perceptual Loss for Robust Unsupervised Homography Estimation}

\author{Daniel Koguciuk, Elahe Arani, Bahram Zonooz \\
Advanced Research Lab, NavInfo Europe, Eindhoven, The Netherlands \\
{\tt\small \{daniel.koguciuk, elahe.arani\}@navinfo.eu}, {\tt\small bahram.zonooz@gmail.com} \\
}

\maketitle
\thispagestyle{empty} 


\begin{abstract}
Homography estimation is often an indispensable step in many computer vision tasks. The existing approaches, however, are not robust to illumination and/or larger viewpoint changes. In this paper, we propose bidirectional implicit Homography Estimation (biHomE) loss for unsupervised homography estimation. biHomE minimizes the distance in the feature space between the warped image from the source viewpoint and the corresponding image from the target viewpoint. Since we use a fixed pre-trained feature extractor and the only learnable component of our framework is the homography network, we effectively decouple the homography estimation from representation learning. We use an additional photometric distortion step in the synthetic COCO dataset generation to better represent the illumination variation of the real-world scenarios. We show that biHomE achieves state-of-the-art performance on synthetic COCO dataset, which is also comparable or better compared to supervised approaches. Furthermore, the empirical results demonstrate the robustness of our approach to illumination variation compared to existing methods.

\end{abstract}

\input{figure_tikz}


\section{Introduction}




Given a pinhole camera model assumption, a homography relates any two images of the same planar surface in space or any two images produced by pure rotational movement of the camera~\cite{hartley2004multipleview}. One image, called a source, can be transformed by a $3 \times 3$ homography matrix $H$ as if viewed from the viewpoint of the other image, called target.
Even if the primary assumptions are violated, the homography can be applied as an initial alignment step in other tasks such as mesh flow~\cite{liu2016meshflow} and optical flow~\cite{ilg2017flownet}.
Therefore, homography has been widely used in many computer vision applications such as SLAM~\cite{engel2014lsd, mur2015orb}, image stitching~\cite{szeliski2006image, guo2016joint}, and change detection and description~\cite{buffington2011homography, van2016hierarchical}.

Traditionally, homography estimation was performed using non-learnable approaches either in the pixel-space (direct methods) or in hand-crafted feature space (feature-based methods) \cite{szeliski2006image}. Recently, with the advancement in Deep Neural Networks (DNNs), DeTone \etal~\cite{detone2016deep} proposed a simple CNN architecture trained in an end-to-end fashion. The idea was to directly regress the parameters of a homography and it achieved similar performance to traditional feature-based methods. A more effective approach is presented by Zeng \etal~\cite{zeng2018rethinking}, where the problem was formulated as per-pixel offset regression. However, supervised methods are often unlikely to be used in real-life scenarios, where ground truth homography labeling is prohibitively expensive.


To mitigate this issue, Nguyen \etal~\cite{nguyen2018unsupervised} introduced an end-to-end unsupervised approach. First, a homography is estimated using both input images and then homography estimation is learned by comparing the per-pixel intensity of warped source image and target image. Models trained using this formulation perform surprisingly well even for images with big viewpoint differences. However, they are not robust to big illumination changes and cannot be used in real-life scenarios \cite{johnson2016perceptual, park2017illumination}. In contrast, Zhang \etal~\cite{zhang2019content}, by learning the feature representation used for both homography estimation and image comparison, achieved robustness to different lighting conditions, but not for images with big viewpoint changes. Therefore, robust unsupervised homography estimation for both big illumination and viewpoint changes at the same time is still an open problem.





Instead of learning a shared representation for homography estimation and image comparison \cite{zhang2019content}, we propose to decouple those two tasks by using a dedicated \textit{Loss Network} to compare warped source image and target image. It has been shown that a pre-trained and fixed convolutional neural network as a \textit{Loss Network} can capture perceptual differences \cite{johnson2016perceptual, wang2018esrgan}. The homography is learned implicitly by comparing the images in feature space produced by the \textit{Loss Network}. Since the homography should be invertible, we compare warped source and target images and swap their roles. Therefore, we call our loss function  bidirectional implicit Homography Estimation (\textit{biHomE}) loss.

We exhibit the simplicity and effectiveness of \textit{biHomE} loss by applying it on top of three homography estimation architectures on the Synthetic COCO (S-COCO) dataset~\cite{detone2016deep}. \textit{biHomE} not only outperforms other unsupervised methods but is also on par with its supervised counterpart. Next, we use an additional photometric distortion augmentation on synthetic COCO (denoted as PDS-COCO) to mimic the illumination changes in real-life scenarios. We show that \textit{biHomE} is the only unsupervised method that can converge on PDS-COCO, while still maintaining comparable or better performance compared to supervised approaches. 
To further demonstrate the effectiveness of our loss, we provide an experimental study on the influence of image alignment on the quality of generated change captions by the Dual Dynamic Attention Model (DUDA)~\cite{park2019robust-change-captioning} method. We achieve a new state-of-the-art performance on the change captioning task.
Our contributions are as follows:
\begin{itemize}
    \item We introduce a new perceptual loss (\textit{biHomE}) to be used in an unsupervised setting, which decouples homography estimation from representation learning for image comparison.
    \item We propose to use an additional photometric distortion step in the synthetic COCO dataset generation (PDS-COCO) to evaluate the robustness of homography estimation to big illumination and viewpoint changes, which is more aligned with real-life scenarios.
    \item We achieve state-of-the-art performance using \textit{biHomE} loss on both S-COCO and PDS-COCO datasets for unsupervised homography estimation.
    \item We apply \textit{biHomE} for unsupervised image alignment to achieve a state-of-the-art change captioning quality on the CLEVR\_Change dataset~\cite{park2019robust-change-captioning}.
\end{itemize}



\section{Related Work}
\label{sec::homography_estimation}

\subsection{Traditional Approaches}

There are two families of traditional approaches for homography estimation: direct methods and feature-based methods \cite{szeliski2006image}. Direct methods work in pixel intensity space, where most of the studies follow the Lucas-Kanade algorithm \cite{lucas1981iterative}.
There are many improved approaches, including utilizing different error metrics \cite{evangelidis2008parametric} or Fourier domain representation \cite{lucey2012fourier}.
Feature-based approaches mostly consist of three components: keypoint detection (such as SIFT~\cite{lowe2004distinctive} and ORB~\cite{rublee2011orb}), correspondence matching (Euclidean distance, correlation), and homography estimation using Direct Linear Transform (DLT) \cite{hartley2004multipleview} with  RANSAC~\cite{fischler1981random} outlier rejection. Feature-based methods usually perform better than direct methods, however, their success depends on the quality of the hand-crafted stages and the content of the image itself.
Poorly distributed features (\ie repetitive pattern or texture), illumination variations, or large viewpoint differences~\cite{wu2007improved} are challenging for both direct and feature-based methods.


\subsection{Supervised Approaches}

DeTone \etal~\cite{detone2016deep} uses a deep learning model to estimate homography, which comes on par with or better than traditional methods. Input images are concatenated channel-wise and passed through 10-layer VGG-style~\cite{simonyan2014very} CNN with a fully connected layer with eight outputs ($x$-$y$ coordinates for 4 points) on top. The problem is then formulated as a 4-point homography regression given the ground truth transformation. On the other hand, Zeng \etal~\cite{zeng2018rethinking} estimates pixel-to-pixel bijection between two images using a U-Net~\cite{ronneberger2015u} architecture. The final homography during inference is produced using RANSAC and DLT in the postprocessing stage. Their method achieves a performance boost compared to DeTone's \etal~\cite{detone2016deep} approach. 


%

\subsection{Unsupervised Approaches}

Nguyen \etal~\cite{nguyen2018unsupervised} formulated the 4-point homography regression as an unsupervised approach. The idea is to minimize pixel-wise intensity error between the warped source image and the target image. They achieve comparable performance to the supervised DeTone \etal~\cite{detone2016deep} method on the S-COCO dataset. Their method, however, cannot compensate for larger illumination changes between images as the learning is performed using pixel intensity comparison.
Zhang \etal~\cite{zhang2019content}, applies the loss function to the feature space instead of the pixel-space. The authors propose to minimize the feature distance between the warped source and the target images and maximize the feature difference between the source and warped source images. Although presented triple loss is efficient for images with small viewpoint shifts, it is not robust to big viewpoint changes.

\section{Method}

\begin{figure}[t]
    \centering
	\begin{tikzpicture}[image/.style = {inner sep=0pt, outer sep=0pt}, node distance = 5mm] 
	\node [image] (frame11)
	{\includegraphics[width=0.49\columnwidth]{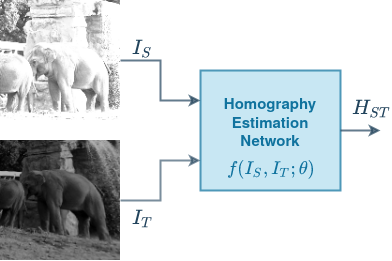}};
 	\node [image,right=of frame11] (frame12) 
 	{\includegraphics[width=0.3\columnwidth]{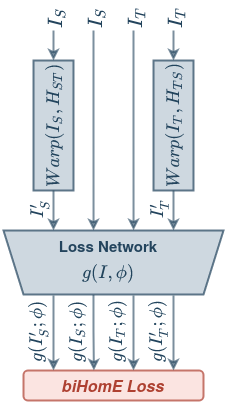}};
 	\end{tikzpicture}

    \caption{Proposed bidirectional implicit Homography Estimation (\textit{biHomE}) loss, where $f$ is any architecture producing homography $H_{ST}$ between the input images $I_S$ and $I_T$ and $g$ is a pre-trained and frozen \textit{Loss Network}. In our experiments, we use ResNet34~\cite{he2016resnet} as feature extractor $g$.}
    \label{fig::ihome-general}
\end{figure}

There are two key insights of our method. The first one is that splitting the architecture into two separate components effectively decouples the homography estimation from representation learning. The second insight of our method is that we want to leverage the fact that convolutional neural networks pre-trained for image classification have already learned to encode meaningful information, which can be used in downstream tasks like image comparison for homography estimation assessment.  Therefore, as shown in Figure \ref{fig::ihome-general}, the system is composed of two components:  The \textit{Homography Estimation Network} (\textit{HEN}) $f$, which can be realized by any architecture able to produce a transformation matrix $H$ and a fixed \textit{Loss Network} $g(I; \phi)$ that is used to define loss function $L$. Homography estimation is learned implicitly by minimizing the loss function defined in the feature space produced by \textit{Loss Network} $g$.

\subsection{Homography Estimation Networks}

The goal of \textit{HEN} is to estimate a $3 \times 3$ homography matrix between two images $I_S$ and $I_T$ using the learnable parameters $\theta$:
\begin{equation} \label{eq:homo}
    H_{ST} = f(I_S, I_T; \theta)
\end{equation}
\noindent Modern \textit{HENs} can be divided into two categories: architectures that directly produce homography (DeTone \etal~\cite{detone2016deep} and Zhang \etal \cite{zhang2019content}) and models that produce per pixel offset (Zeng \etal \cite{zeng2018rethinking}). The former can be directly applied into our framework, but that is not the case with the latter. Thus, we first randomly sample without replacement $N_S < H \cdot W$ points out of the output perspective field of size $H \times W$ and then use Direct Linear Transform (DLT) \cite{hartley2004multipleview} to estimate the homography matrix in an end-to-end pipeline.

\subsection{Perceptual Loss Functions}

In the second part of the pipeline, homography $H_{ST}$ estimated by \textit{HEN} is used to transform source image $I_S$ into $I'_S = Warp(I_S, H_{ST})$. Rather than encouraging the pixels of the warped source image $I'_S$ to match the pixels of the target image $I_T$ (similar to \cite{nguyen2018unsupervised}), 
we rely on perceptual similarity of high-level features extracted from pre-trained \textit{Loss Network} $g$.
Let $g_{j}(I)$ be the feature map of size $C_j \times H_j \times W_j$ after $j$th layer of the \textit{Loss Network} $g$ for a given image $I$. We define the mask $M$ of the same size as $g_{j}(I)$ consisting of all ones and use $M' = Warp(M, H_{ST})$ to make sure the loss is applied only to the part visible on both images. The perceptual loss can be defined in multiple ways.


\noindent \textbf{Implicit Homography Estimation with MAE/MSE Loss.} The simplest approach is to define loss as per pixel normalized distance in feature space produced by $g$ between warped source image $I'_S$ and target image $I_T$:
\begin{equation}\label{eq:loss1}
\begin{split}
    L_{d}(I_S, I_T) &= \frac{\sum\limits_i^{H_{j},W_{j},C_{j}} M'_i  \cdot d(g_{ji}(I_S'), g_{ji}(I_T))}{\sum\limits_i^{H_{j},W_{j},C_{j}} M'_i}
\end{split}
\end{equation}
\noindent where $d$ is either $L_1$ distance or $L_2$ distance squared, resulting in masked $MAE$ or masked $MSE$ perceptual loss functions.


\noindent \textbf{Implicit Homography Estimation Loss (\textit{iHomE}).} Rather than directly minimizing only the distance between $I_T$ and $I'_S$, we can encourage the network to simultaneously push $I_T$ away from the original $I_S$ in triplet fashion. The loss is highly inspired by Zhang \etal \cite{zhang2019content}, but instead of learning the feature space to compare images, we use a fixed pre-trained \textit{Loss Network} and non-learnable mask $M_S$. The iHomE loss can be formulated as:
\begin{equation}\label{eq:loss_ihome_agn}
\begin{split}
    L_{iHomE}(I_S, I_T) &= \frac{\sum\limits_i^{H_j,W_j} M'_{i} \cdot \max (ap_{ji} - an_{ji} + m, 0)}{\sum\limits_i^{H_j,W_j} M'_{i}}
\end{split}
\end{equation}
\noindent where $m$ is triplet loss margin and $ap_{ji}$ and $an_{ji}$ are anchor-positive and anchor-negative channel-aggregated distances:
\begin{equation}\label{eq:channel_aggregation_agn}
\begin{split}
    ap_{ji} &= \frac{1}{C} \sum\limits_k^{C_{j}} ||g_{jik}(I_T) - g_{jik}(I_S')||_1 \\
    an_{ji} &= \frac{1}{C} \sum\limits_k^{C_{j}} ||g_{jik}(I_T) - g_{jik}(I_S)||_1
\end{split}
\end{equation}


\noindent \textbf{Bidirectional implicit Homography Estimation Loss (\textit{biHomE}).} We can additionally leverage the fact that homography is invertible by estimating the transformation from the target $I_T$ to the source $I_S$ images. Following Zhang \etal \cite{zhang2019content}, we also add a constraint that enforces $H_{ST}$ and $H_{TS}$ to be inverse. We can formulate \textit{biHomE} as:
\begin{equation}
\begin{split}
    L_{biHomE}(I_S, I_T) = & L_{iHomE}(I_S, I_T) + \\
                             & L_{iHomE}(I_T, I_S) + \\
                             & \mu ||H_{ST}H_{TS} - \mathbb{1}||^2_2
\end{split}
\end{equation}
\noindent where $\mu$ is the balancing hyperparameter \cite{zhang2019content} 
and $\mathbb{1}$ is $3 \times 3$ identity matrix.


\section{Datasets}

Since collecting the data for homography learning along with ground truth is hard, DeTone \etal \cite{detone2016deep} proposed to generate the dataset applying random projective transformations to COCO \cite{lin2014microsoft}. The Synthetic COCO (S-COCO) dataset was accepted by the research community, however, it does not model photometric  distortion  present  in  real-world  images, such as contrast, brightness, and saturation changes. Here, we introduce a Photometrically Distorted Synthetic COCO (PDS-COCO) where we artificially model illumination changes by utilizing photometric distortion techniques used in \cite{liu2016ssd} as augmentation for SSD object detector.


\subsection{Synthetic COCO (S-COCO)}
\label{sec::scoco}

\begin{figure}[t]

	\centering
	\resizebox{!}{0.4\textheight}{
	\begin{tikzpicture}[
	image/.style = {text width=0.33\columnwidth, 
		inner sep=0pt, outer sep=0pt},
	node distance = 1mm and 1mm,
	line width=0.5mm
	] 
	
	\node [image] (coco)
	{\includegraphics[width=\linewidth]{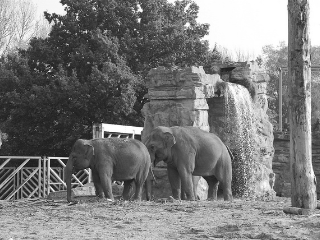}};
  	\node[rotate=90,left=of coco,anchor=south,yshift=-2mm] {coco\_image};

    
	\node [image,below left=of coco,xshift=+5mm,yshift=-8mm] (image1) 
	{\includegraphics[width=\linewidth]{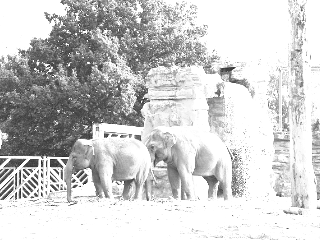}};
  	\node[rotate=90,left=of image1,anchor=south,yshift=-1mm] {source-candidate};
  	\draw (coco.south) edge[->] node[sloped,anchor=center,xshift=0mm,yshift=2mm] {distort / copy} (image1.north);

    \node [image,below=of image1,xshift=0mm,yshift=-8mm] (image1rect) 
	{\includegraphics[width=\linewidth]{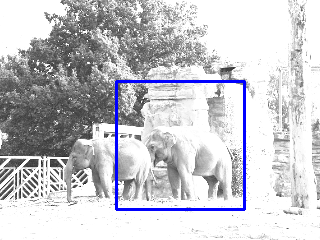}};
  	\draw (image1.south) edge[->] node[sloped,anchor=center,yshift=-2mm,rotate=180] {crop} (image1rect.north);
  	

	\node [image,below right=of coco,xshift=-5mm,yshift=-8mm] (image2) 
	{\includegraphics[width=\linewidth]{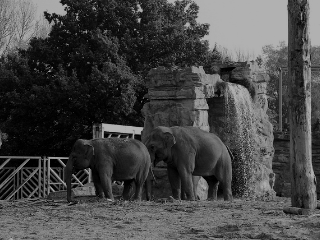}};
  	\node[rotate=90,left=of image2,anchor=south,yshift=-2mm] {target-candidate};
  	\draw (coco.south) edge[->] node[sloped,anchor=center,xshift=2mm,yshift=2mm] {distort / copy} (image2.north);

    \node [image,below=of image2,xshift=0mm,yshift=-8mm] (image2rect) 
	{\includegraphics[width=\linewidth]{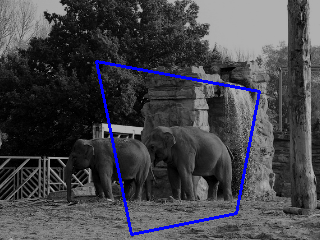}};
	\draw (image2.south) edge[->] node {} (image2rect.north);
  	\draw (image1rect.east) edge[->] node[anchor=center,yshift=2mm] {perturb} (image2rect.west);
	
	\node [image,below=of image2rect,xshift=0mm,yshift=-8mm] (image2recthba) 
	{\includegraphics[width=\linewidth]{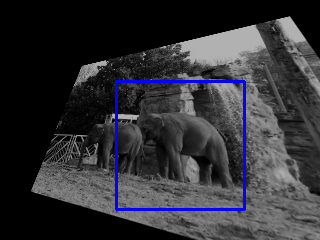}};
	\draw (image2rect.south) edge[->] node[sloped,anchor=center,yshift=-2mm,rotate=180] {$H_{TS}$} (image2recthba.north);
	

	\node [image,below left=of image2recthba,xshift=-9mm,yshift=-8mm,style={text width=0.15\columnwidth}] (patch1) 
	{\includegraphics[width=\linewidth]{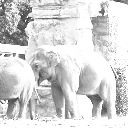}};
	\node[below=of patch1,yshift=1mm] {$I_S$};

    \node [image,below left=of image2recthba,xshift=5mm,yshift=-8mm,style={text width=0.15\columnwidth}] (patch2)
	{\includegraphics[width=\linewidth]{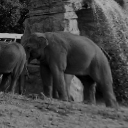}};
  	\draw (image2recthba.south) edge[->,bend left=10] node[] {} ($(patch2.north)+(0mm,2mm)$);
  	\draw (image1rect.south) edge[->,bend right=10] node[] {} ($(patch1.north)+(0mm,2mm)$);
  	\node[below=of patch2,yshift=1mm] {$I_T$};
	
 	\end{tikzpicture}}

    \caption{PDS-COCO dataset generation. Two separate photometric distortions are performed for a given COCO image resulting in the creation of source-candidate and target-candidate. Next, a random crop is selected for the source-candidate. The random crop is perturbed and given these correspondences, $H_{ST}$ is computed. $H_{TS} = (H_{ST})^{-1}$ is applied to the target-candidate. Finally, source image $I_S$ and target image $I_T$ are extracted. Substituting photometric distortion with copy operation gives the original S-COCO dataset generation procedure.}
    \label{fig::appendix-pds-coco}
\end{figure}

The dataset was introduced by DeTone \etal~\cite{detone2016deep} in which source and target candidates are generated by duplicating the same COCO image~\cite{lin2014microsoft}. The source patch $I_S$ is generated by randomly cropping a source candidate at position $p$ with a size of $128 \times 128$ pixels. Then the patch's corners are randomly perturbed vertically and horizontally by values within the range $[-\rho, \rho]$ and the four correspondences define a homography $H_{ST}$. The inverse of this homography $H_{TS} = (H_{ST})^{-1}$ is applied to the target candidate and from the resulted warped image a target patch $I_T$ is cropped at the same location $p$. Both $I_S$ and $I_T$ are the input data with the homography $H_{ST}$ as ground truth as shown in Figure~\ref{fig::appendix-pds-coco}. Such a procedure not only allows us to use large scale datasets but also creates ground truth homography labels for each transformation. 

\begin{figure*}[t]
    \centering
    
    
    \includegraphics[height=0.139\textheight]{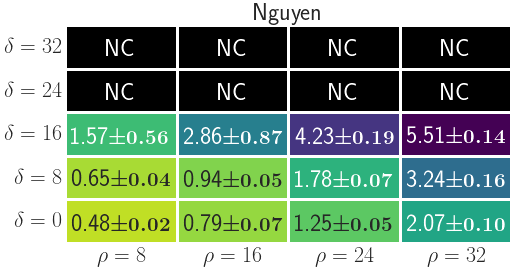}
    \includegraphics[height=0.139\textheight]{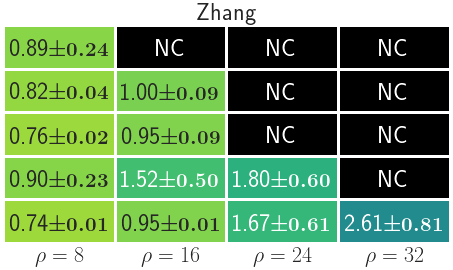}
    \includegraphics[height=0.139\textheight]{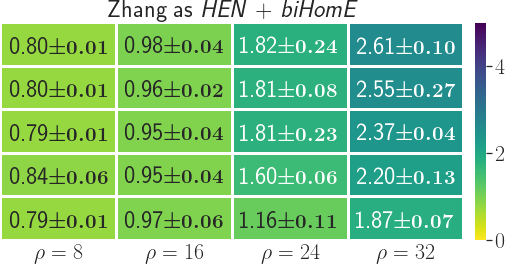}
    
    
    \caption{The viewpoint and illumination robustness comparisons of three unsupervised methods for different levels of viewpoint $\rho$ and illumination $\delta$ changes. S-COCO is equivalent to the box with $\rho=32$ and $\delta=0$ and PDS-COCO is equivalent to the box with $\rho=32$ and $\delta=32$. NC stands for Not Converged (the methods which were not able to converge at least once per twenty independent runs). The method of Nguyen \etal \cite{nguyen2018unsupervised} is robust to big viewpoint changes (big values of $\rho$) but is not able to converge for big illumination distortion (big values of $\delta$). Zhang \etal \cite{zhang2019content} method is highly robust to illumination change $\delta$ but only for small viewpoint change $\rho$, whereas our \textit{biHomE} loss is robust to both big illumination and viewpoint changes.}
    \label{fig::illumination-robustness}
\end{figure*}


\subsection{Photometrically Distorted Synthetic COCO (PDS-COCO)}
\label{sec::pdscoco}

To test the robustness of different homography estimation architectures to illumination changes we introduce Photometrically Distorted Synthetic COCO (PDS-COCO) dataset. The photometric distortion is inspired by Liu \etal~\cite{liu2016ssd} and the first step involves adjusting the brightness of the image using randomly picked value $\delta_b \in \mathcal{U}(-32, 32)$. Next, contrast, saturation and hue noise is applied with the following values: $\delta_c \in \mathcal{U}(0.5, 1.5)$, $\delta_s \in \mathcal{U}(0.5, 1.5)$ and $\delta_h \in \mathcal{U}(-18, 18)$. Finally, the color channels of the image are randomly swapped with a probability of $0.5$. A composition of these distortions produces vastly different images, which constitutes a challenge for current homography estimation architectures. Such a photometric distortion procedure is applied to the original image independently to create source and target candidates. The rest of the procedure is the same as for S-COCO dataset generation as shown in the Figure \ref{fig::appendix-pds-coco}.


\section{Experiments}
\label{sec::experiments}

\subsection{Experimental Setup}

We implemented all the methods in PyTorch~\cite{NEURIPS2019_9015} and Kornia \cite{riba2020kornia}. We modified the \textit{HEN} architectures (if necessary) to use a ResNet34-like \cite{he2016resnet} feature extractor, to make sure the performance difference between methods comes from better design instead of a better CNN backbone. All methods are trained using the Adam optimizer \cite{kingma2014adamoptimizer} with a batch size of 64 and for $90k$ iterations. The initial learning rate (Detone and Nguyen: $5e^{-3}$, Zeng: $1e^{-3}$, Zhang: $1e^{-2}$) is decayed by a factor of 10 every $30k$ iterations. We use ResNet34 as a fixed \textit{Loss Network} $g$ with features taken after the first residual block. We also used $N_S = 1024$ for Zeng method and $\mu = 0.01$, unless stated otherwise. 
The homography estimation quality is reported using Mean Absolute Corner Error (MACE) \cite{detone2016deep}. We report the mean and standard deviation of three runs.

\subsection{S-COCO and PDS-COCO Results}
\label{sec::scoco_results}

Our implementation achieves much lower MACE on S-COCO than originally reported by DeTone \etal~\cite{detone2016deep} and Nguyen \etal \cite{nguyen2018unsupervised}: $1.96$ and $2.07$ against $9.20$ and $12.91$, respectively. This is because we employ a modern CNN architecture -- ResNet34 \cite{he2016resnet} instead of originally used VGG-like structure \cite{simonyan2014very}. Having the same backbone, Zeng's \etal \cite{zeng2018rethinking} model is only about $13\%$ better than DeTone's \etal \cite{detone2016deep} model -- instead of $564\%$ improvement reported in their paper. 

As shown in the Figure \ref{fig::exp-1}, our architecture gives the best performance compared to all other unsupervised methods. Specifically, using our \textit{biHomE} loss with the same \textit{HEN} as Zhang \etal \cite{zhang2019content} demonstrates the beneficial effect of disentangling the homography estimation learning from feature representation learning. Moreover, our \textit{biHomE} loss achieves comparable (Zeng \etal \cite{zeng2018rethinking}) or better (DeTone \etal \cite{detone2016deep}) homography estimation performance than supervised methods on S-COCO. 




Homography estimation on PDS-COCO presents a bigger challenge, because of additional photometric distortions on the images. Supervised approaches perform much worse in terms of MACE and neither Nguyen's~\cite{nguyen2018unsupervised} nor Zhang's~\cite{zhang2019content} methods can converge. Our \textit{biHomE} loss is the only method trained in an unsupervised manner that can converge on PDS-COCO. Similar to S-COCO, our \textit{biHomE} loss achieves comparable or better performance compared to supervised methods.



\begin{table*}[t]
\centering
\resizebox{\textwidth}{!}{
\begin{tabular}{@{}c|c|ccc|c|c|c|c@{}}
\toprule
\multicolumn{1}{l|}{\textbf{}} & \textbf{\#1}        & \textbf{\#2}                             & \textbf{\#3}                             & \textbf{\#4}                             & \textbf{\#5}                          & \textbf{\#6}                       & \textbf{\#7}                          & \textbf{\#8}                           \\ \midrule
\textbf{Loss network}          & VGG19               & {\color[HTML]{FE0000} \textbf{ResNet34}} & {\color[HTML]{FE0000} \textbf{ResNet34}} & {\color[HTML]{FE0000} \textbf{ResNet34}} & ResNet34                              & ResNet34                           & ResNet34                              & ResNet34                               \\
\textbf{Layers}                & VGG22               & {\color[HTML]{000000} Block2}            & {\color[HTML]{FE0000} \textbf{Block1}}   & Block0                                   & Block1                                & Block1                             & Block1                                & Block1                                 \\
\textbf{Activation}            & Before              & Before                                   & {\color[HTML]{000000} Before}            & Before                                   & {\color[HTML]{FE0000} \textbf{After}} & After                              & After                                 & After                                  \\
\textbf{Loss function}         & MSE                 & MSE                                      & MSE                                      & {\color[HTML]{000000} MSE}               & {\color[HTML]{000000} MSE}            & {\color[HTML]{FE0000} $\boldsymbol{L_1}$} & {\color[HTML]{FE0000} \textbf{iHomE}} & {\color[HTML]{FE0000} \textbf{biHomE}} \\ \midrule
\textbf{MACE}                  
& $\boldsymbol{8.59}$ \scriptsize{$\pm 0.05$}
& $7.15$ \scriptsize{$\pm 0.16$}                              
& $\boldsymbol{4.18}$ \scriptsize{$\pm 0.16$}
& $4.74$ \scriptsize{$\pm 0.05$}
& $\boldsymbol{4.11}$ \scriptsize{$\pm 0.11$}
& $\boldsymbol{3.82}$ \scriptsize{$\pm 0.11$}
& $\boldsymbol{3.67}$ \scriptsize{$\pm 0.10$}
& $\boldsymbol{2.39}$ \scriptsize{$\pm 0.20$}                    \\ \bottomrule
\end{tabular}}
\caption{Comparison of performance on PDS-COCO of different settings of perceptual loss. Each column represents a model with DeTone's \cite{detone2016deep} architecture as HEN and a particular perceptual loss configuration. The red color indicates the main improvement compared with the previous model starting from Wang \etal \cite{wang2018esrgan} setting.}
\label{tab::ablation}
\end{table*}


\subsection{Illumination and Viewpoint Robustness Study}
\label{sec::illumination_robustness}

The real-life images can exhibit big illumination and viewpoint changes. It is hard to collect such datasets with ground truth homography labels, however, we can use the COCO dataset and simulate different conditions by two parameters: viewpoint change $\rho$ and illumination change $\delta$.

Viewpoint change $\rho$ defines the maximum range of corner perturbation, and it was already introduced in Section \ref{sec::scoco}. Since photometric distortion has a few elements, we bring them all together by illumination change $\delta$ so that all other deltas are controlled by one parameter:
\begin{equation}\label{eq:illumination_robustness}
\begin{split}
\delta_x \in \mathcal{U}(-X \cdot \frac{\delta}{32}, X \cdot \frac{\delta}{32})
\end{split}
\end{equation}
\noindent where $\delta_x$ is one of $\delta_b, \delta_c, \delta_s$ or $\delta_h$ (following the notation from Section \ref{sec::scoco}), and X is the corresponding perturbation range. The bigger $\rho$ the more viewpoint change and respectively the bigger $\delta$ the more illumination change between input images. For such a notation \textit{S-COCO} is produced by $\rho=32$ and $\delta=0$ and \textit{PDS-COCO} is produced by $\rho=32$ and $\delta=32$.


As presented in Figure \ref{fig::illumination-robustness}, the method of Nguyen \etal \cite{nguyen2018unsupervised} based on photometric loss is robust to big viewpoint changes (big values of $\rho$), but is not able to produce any reasonable model for big illumination distortion $\delta$. On the other hand, Zhang \etal \cite{zhang2019content} method is highly robust to illumination change $\delta$ but only for small viewpoint change $\rho$. Our \textit{biHomE} loss, which for a fair comparison we add on top of the same \textit{HEN} as Zhang \etal \cite{zhang2019content} is robust both to big illumination $\delta$ and viewpoint $\rho$ changes.

\begin{table}[b]
\centering
\begin{tabular}{@{}lccc@{}}
\toprule
\textit{HEN}  & Zhang        & Zhang           & Zhang            \\ 
Loss Function & Triplet Loss & \textit{biHomE} & \textit{biHomE}  \\
Channel-      & -            & c-wise           & c-agnostic \\ \midrule
$m=1$   & $3.07$ \scriptsize{$\pm 0.34$} & $4.57$ \scriptsize{$\pm 0.21$}                                                            &  $3.33$ \scriptsize{$\pm 0.34$}                                                                       \\
$m=100$  & $2.50$ \scriptsize{$\pm 0.25$}           & $2.84$ \scriptsize{$\pm 0.27$}                                                            &  $1.86$ \scriptsize{$\pm 0.16$}                                                            \\
$m=\infty$ & $\boldsymbol{2.08}$ \scriptsize{$\pm0.11$} & $\boldsymbol{1.87}$ \scriptsize{$\pm 0.07$} & $\boldsymbol{1.87}$ \scriptsize{$\pm 0.07$}                                     \\ \bottomrule
\end{tabular}
\caption{The performance on the S-COCO dataset of Zhang \etal \cite{zhang2019content} and our \textit{biHomE} loss with different margin $m$ settings and different channel information aggregation. The \textit{infinite} margin version is consistently better for both losses and it is used in all experiments in the paper.}
\label{tab::margin}
\end{table}


\subsection{From Perceptual Loss to \textit{biHomE}}

Perceptual Loss was already introduced in the tasks of super-resolution \cite{bruna2016super, ledig2017photo, wang2018esrgan}, style transfer \cite{johnson2016perceptual}, image denoising \cite{gholizadeh2019deep}, or training autoencoders \cite{pihlgren2020improving}. The idea was to rely on perceptual similarity of high-level features extracted from pre-trained networks. The same property is desired in homography estimation. The architecture of \textit{Loss Network} designed for super-resolution task, however, could be not optimal in the task of image comparison for homography estimation. To study the effects of different components of the perceptual loss we gradually modified the State-of-the-Art perceptual loss proposed by Wang \etal \cite{wang2018esrgan} and compare their performance in the homography estimation task as presented in Table \ref{tab::ablation}. The red color indicates the main improvement compared with the previous model. We use DeTone \etal \cite{detone2016deep} in all configurations and only change elements of perceptual loss. The average MACE of three runs is presented. A detailed discussion is provided as follows.

\noindent \textbf{ResNet34.} We first move from VGG19 \cite{simonyan2014very} network to ResNet34 \cite{he2016resnet}. Using ResNet as a \textit{Loss Network} significantly improves the homography estimation accuracy, which can be caused by a better-learned feature representation. We also noticed that learning is more likely to converge with ResNet, probably due to better gradient flow of residual connections. The detailed comparison is out of the scope of this article. For a comprehensive analysis of these networks please refer to \cite{bianco2018benchmark, khan2020survey}.

\noindent \textbf{Block1.} Ledig \etal \cite{ledig2017photo} argues to use higher-level features from deeper network layers with the potential to focus more on the content of the image. It is essential for the Super-Resolution task, but as depicted in Table \ref{tab::ablation}, in Homography Estimation low-level features matter more.

\noindent \textbf{Features after activation.} In contrast to Wang \etal \cite{wang2018esrgan}, we show that it is beneficial to use features after the ReLU activation function \cite{glorot2011deep}. A possible explanation is that low-level features are activated mostly on edges and corners, which are crucial for image alignment purposes (please refer to Figure 6a of Wang \etal \cite{wang2018esrgan} for a visual comparison).

\noindent $\boldsymbol{L_1}$ \textbf{distance.} Replacing $MSE$ loss function with $L_1$ distance yields better homography estimates. Due to warping and illumination differences, feature maps could contain some artifacts, and $L_1$ loss is known to be more robust to outliers~\cite{goodfellow2016deep}.

\noindent \textbf{iHomE.} Triplet loss introduced by Zhang \etal \cite{zhang2019content} can further improve the homography estimation performance. This is probably because the model has additional information of original $I1$ image features. 

\noindent \textbf{biHomE.} Using triplet loss in both directions similarly to Zhang \etal \cite{zhang2019content} can additionally improve the results. The improvement is even more profound for other \textit{HENs}.


\subsection{Margin Ablation}

Zhang \etal \cite{zhang2019content} used a margin value of $m=1$ in their triplet loss formulation. However, we found out that on S-COCO bigger values of $m$ yield better results (Table \ref{tab::margin}). The best performance is produced when the triplet loss is never saturated, which is equivalent to setting the margin to an \textit{infinite} value:
\begin{equation}\label{eq:loss_ihome_inf}
\begin{split}
    L_j^{iHomE}(I_S, I_T) &= \frac{\sum\limits_i^{H_{j}W_{j}} M'_i \cdot (ap_{ji} - an_{ji})}{\sum\limits_i^{H_{j}W_{j}} M'_i}
\end{split}
\end{equation}
\noindent In all the experiments for both Zhang \etal \cite{zhang2019content} and our iHomE and \textit{biHomE} losses we use above Formulation \ref{eq:loss_ihome_inf} instead of Equation \ref{eq:loss_ihome_agn}.


\subsection{Channel Information Aggregation}

In contrast to Zhang \etal \cite{zhang2019content}, our feature map obtained by \textit{Loss Network} $g$ has more than one channel, thus we have to figure out how the channel information will be aggregated. One can either calculate distances for every channel and then apply triplet loss in a channel-agnostic way per every spatial location or apply triplet per every channel location (channel-wise). The former is presented in equations \ref{eq:loss_ihome_agn}~and~\ref{eq:channel_aggregation_agn}, and the latter can be formulated as: 
\begin{equation}\label{eq:loss_ihome_wise}
\begin{split}
    L_j^{iHomE}(I_S, I_T) &= \frac{\sum\limits_{i}^{H_{j}W_{j}C_{j}} M'_{i} \cdot \max (ap_{ji} - an_{ji} + m, 0)}{\sum\limits_i^{H_{j}W_{j}C_{j}} M'_{i}}
\end{split}
\end{equation}
\noindent where;
\begin{equation}\label{eq:channel_aggregation_wise}
\begin{split}
    ap_{ji} &= ||g_{ji}(I_T) - g_{ji}(I_S')||_1 \\
    an_{ji} &= ||g_{ji}(I_T) - g_{ji}(I_S)||_1
\end{split}
\end{equation}

As presented in Table \ref{tab::margin} the channel-agnostic way achieves better performance on the S-COCO dataset. We hypothesize that forcing every channel of $I_T$ feature description to be closer to $I_S'$ than to $I_S$ is probably a very hard task and looking at all channels before comparison makes the optimization easier. Channel-agnostic and channel-wise versions of \textit{biHomE} loss with \textit{infinite} margin are mathematically equivalent, so we report only one number.


\subsection{To Freeze or not to Freeze}

We also want to find out how important is freezing the \textit{Loss Network} for effective homography estimation learning. As shown in Table \ref{tab::unfix}, fixed \textit{Loss Network} performs better. One of the possible reasons is that freezing weights of $g$ allows using a bigger learning rate, which can result in better convergence \cite{smith2018disciplined, smith2019super}. But even for the same learning rate value freezing the \textit{Loss Network} is still a preferable policy.


\begin{table}[h]
\centering
\begin{tabular}{@{}lcc@{}}
\toprule
                       & lr & S-COCO MACE \\ \midrule
Zhang+\textit{biHomE}           & 1e-2          & $\boldsymbol{1.87}$ \scriptsize{$\pm 0.07$}  \\
Zhang+\textit{biHomE}           & 1e-3          & $2.03$ \scriptsize{$\pm 0.04$} \\
Zhang+\textit{biHomE} (learnable) & 1e-3          & $2.17$ \scriptsize{$\pm 0.13$} \\ \bottomrule
\end{tabular}
\caption{S-COCO performance of Zhang \etal \cite{zhang2019content} as \textit{HEN} and our \textit{biHomE} loss when using fixed and learnable \textit{Loss Network}. Fixing weights of $g$ allows using bigger learning rate, which can result in better convergence \cite{smith2018disciplined, smith2019super}.}
\label{tab::unfix}
\end{table}


\subsection{\textbf{\textit{biHomE}} Performance on Out-of-Distribution Dataset}

One of the shortcomings of using models pre-trained on ImageNet \cite{russakovsky2014imagenet} as \textit{Loss Network} is possible sub-optimal results on out-of-distribution datasets. To test this vulnerability we compare Zhang \etal \cite{zhang2019content} method with the original and our \textit{biHomE} loss on FLIR Dataset \cite{flirdataset} preprocessed similar to S-COCO. As shown in Table \ref{tab::flir}, although thermal images are vastly different than color images the \textit{Loss Network} was trained on, our \textit{biHomE} loss still achieves better performance.

\begin{table}[h]
\centering
\begin{tabular}{@{}lc@{}}
\toprule
                       & S-FLIR MACE \\ \midrule
Zhang+\textit{biHomE} & $\boldsymbol{1.59}$ \scriptsize{$\pm 0.13$} \\
Zhang+TripletLoss           & $2.08$ \scriptsize{$\pm 0.25$} \\ \bottomrule
\end{tabular}
\caption{The performance of our \textit{biHomE} loss vs. original Zhang \etal \cite{zhang2019content} method on S-FLIR dataset \cite{flirdataset} containing thermal images for ADAS systems.}
\label{tab::flir}
\end{table}


\section{Image Alignment for Change Captioning}

In this Section, we study the influence of image alignment using homography on change captioning task. The goal is to generate captions for an image pair before and after the change. There are six change types (color/material change, adding/dropping/moving an object, and no change) and each is combined with illumination/viewpoint change. Recently, Park \etal \cite{park2019robust-change-captioning} proposed a new architecture called Dual Dynamic Attention Model (DUDA) for generating change captions. It entails three main components: feature extractor (ResNet101 \cite{he2016resnet}), Dual Attention and Dynamic Speaker modules. The Dual Attention module takes features produced by the feature extractor and generates a~sparse feature vector with a description of the changes between input images. The Dynamic Speaker module then produces the caption word-by-word dynamically moving the attention between the first and second images depending on the stage of the sentence. 


\begin{table*}[t]
\resizebox{\textwidth}{!}{
\begin{tabular}{l|cccc|cccc|cccc}
\hline
                     & \multicolumn{4}{c|}{Total}                            & \multicolumn{4}{c|}{Scene Change}                      & \multicolumn{4}{c}{Distractor}                                 \\
Approach    & B             & C              & M             & S             & B    & C              & M             & S             & B             & C              & M             & S             \\ \cline{1-13} 
\{No alignment\} + DUDA \cite{park2019robust-change-captioning}        & 53.0          & 115.2          & 37.4          & 31.3          & 50.8 & 100.3          & 33.2          & 27.8          & 62.3          & 115.9          & 49.8          & 34.8          \\
\{Zhang \cite{zhang2019content} + TripletLoss\} + DUDA \cite{park2019robust-change-captioning}      & \textbf{54.5} & 120.7          & \textbf{40.2} & \textbf{33.0} & \textbf{52.1} & 111.4          & 36.0          & \textbf{31.4} & 61.2          & 115.3          & \textbf{51.6} & 34.6          \\
\{Zhang \cite{zhang2019content} + \textit{biHomE}\} + DUDA \cite{park2019robust-change-captioning} & 53.0          & \textbf{124.5} & \textbf{40.2} & \textbf{33.0} & 50.6 & \textbf{117.3} & \textbf{36.9} & 30.9          & \textbf{63.7} & \textbf{117.1} & 50.5          & \textbf{35.0} \\ \hline
\end{tabular}}
\caption{Change Captioning evaluation on CLEVR\_Change dataset. Aligning images before learning to generate captions improves all captioning metrics: BLEU-4 (B), CIDEr (C), METEOR (M), and SPICE (S) in each setting (i.e. Total, Scene Change, Distractor). For more details, we refer to Park \etal~\cite{park2019robust-change-captioning}. The model trained with our \textit{biHomE} loss achieves the best result in most of the cases.}
\label{tab::clevr-change}
\end{table*}


\begin{figure}[b]
    \centering
    \includegraphics[width=\columnwidth]{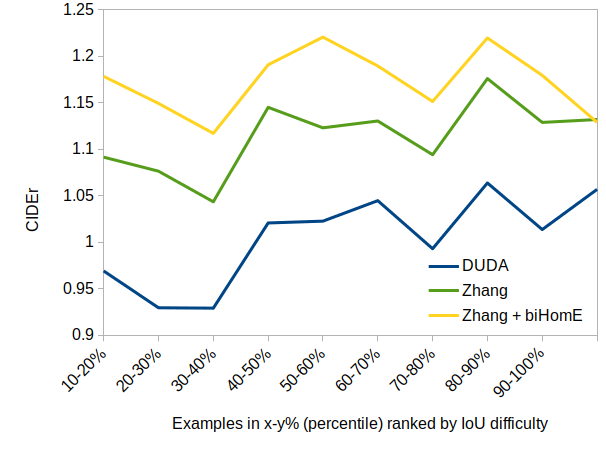}
    \caption{Change captioning performance breakdown by viewpoint shift (measured by IoU). Aligning images with Zhang~\cite{zhang2019content} method before learning to generate captions improves the metrics, but performance degradation between smaller and bigger viewpoint shifts is still present. Learning Zhang~\cite{zhang2019content} with our \textit{biHomE} loss seems to alleviate this problem. 
    }
    \label{fig::cider-by-iou}
\end{figure}


\begin{figure}[b]

	\centering
	\begin{tikzpicture}[
	image/.style = {text width=0.45\columnwidth, 
		inner sep=0pt, outer sep=0pt},
	node distance = 1mm and 1mm
	] 
	
	\node [image] (frame11)
	{\includegraphics[width=\linewidth]{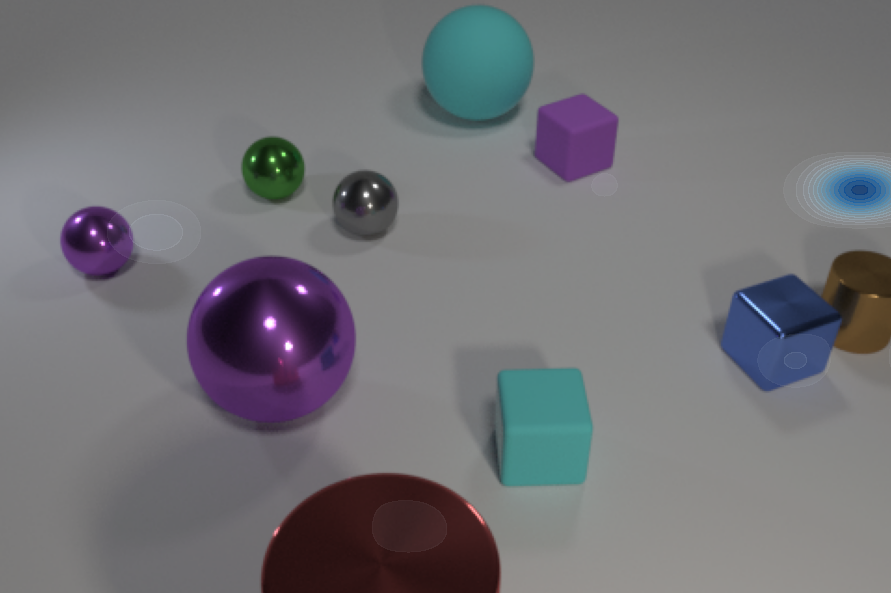}};
	\node [image,right=of frame11] (frame12) 
	{\includegraphics[width=\linewidth]{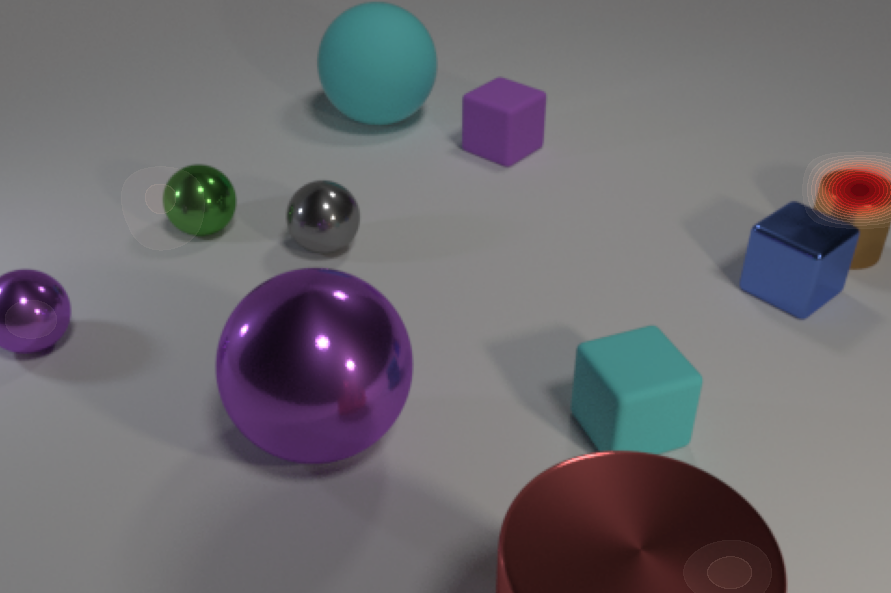}};
	
	\node[below=of frame11,align=left,anchor=west,yshift=-0.4cm,xshift=-1.5cm]  {\small ~~~~the small brown metal cylinder that is in front of \\ \small the small purple rubber object has been newly placed};
 	\draw[dashed,line width=0.05cm] (-1.8cm,-2.25cm) -- (5.75cm,-2.25cm);
	
 	\node [image,below=of frame11,yshift=-1.1cm] (frame21)
 	{\includegraphics[width=\linewidth]{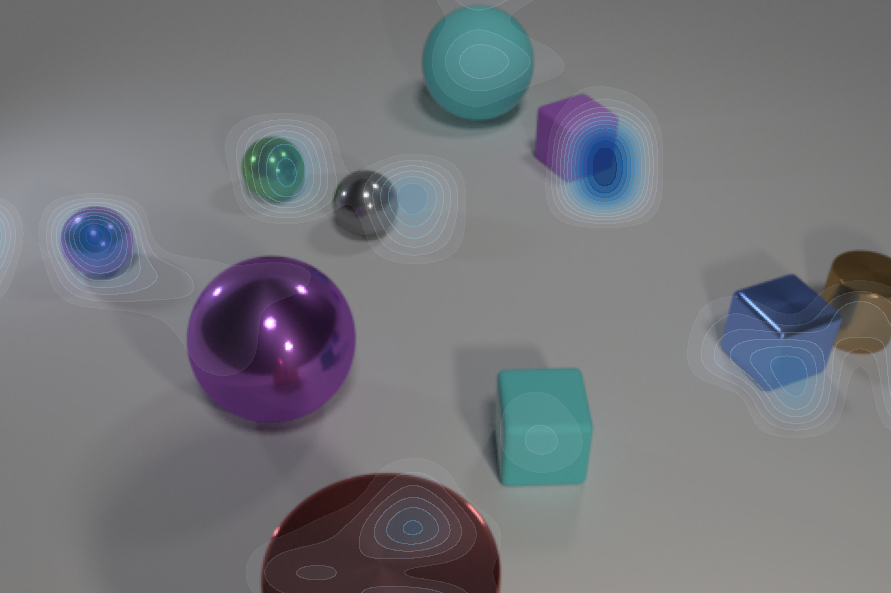}};
 	\node [image,right=of frame21] (frame22) 
 	{\includegraphics[width=\linewidth]{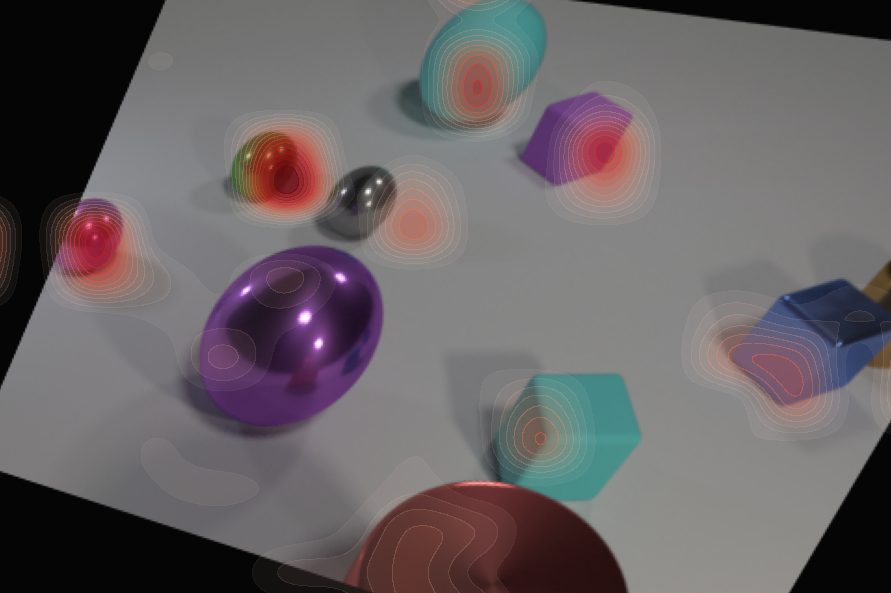}};
	
	\node[below=of frame21,align=left,anchor=west,yshift=-0.1cm,xshift=-0cm]  {\small the scene remains the same};
 	\draw[dashed,line width=0.05cm] (-1.8cm,-2.25cm) -- (5.75cm,-2.25cm);
	
 	\node[above=of frame11] {\small before change};
 	\node[above=of frame12] {\small after change};
	
 	\node[rotate=90,left=of frame11,anchor=south] {\small DUDA};
 	\node[rotate=90,left=of frame21,anchor=south] {\small DUDA + iHomE};
	
 	\end{tikzpicture}

    \caption{Qualitative result comparing DUDA \cite{park2019robust-change-captioning} output with and without image alignment. The blue and red attention maps are applied before and after the change, respectively. After adding the image alignment step, DUDA \cite{park2019robust-change-captioning} architecture produces the correct caption.}
    \label{fig::duda-vis}
\end{figure}

In their evaluation Section, Park \etal \cite{park2019robust-change-captioning} showed degradation of captioning performance as viewpoint shift increases -- our reproduced results are shown in Figure \ref{fig::cider-by-iou}. The Dual Attention module intrinsically assumes the images before and after the change are roughly aligned, thus finding corresponding objects on images with big viewpoint change is challenging. We argue that aligning those images helps DUDA architecture learn better captions of the change on the scene.

We conducted experiments for Zhang \etal \cite{zhang2019content} architecture with and without our \textit{biHomE} loss. Unless stated otherwise, the following are the steps that have been followed in all our experiments: first, the homography estimation is learned. Second, we transform images after change onto images before change using the trained homography estimation model. The pre-processed images are trained using the DUDA model and the captioning quality metric is reported in Table \ref{tab::clevr-change}. Moreover, we also prepare similar validation of the robustness of the model to viewpoint shift, according to the methodology shown in Park \etal \cite{park2019robust-change-captioning} which is depicted in Figure \ref{fig::cider-by-iou}.

Experimental evaluation shows that aligning images before learning to caption is beneficial not only for overall captioning quality but also for performance degradation between smaller and bigger viewpoint shifts. A sample image pair with caption generated with and without image alignment step is presented in Figure \ref{fig::duda-vis}. Note that using our \textit{biHomE} loss the performance degradation is negligible for the CLEVR\_Change dataset, which can be explained by facilitated correspondence search in the Dual Attention module on aligned images.


\section{Conclusion}

We presented an unsupervised approach to homography estimation that is robust to big illumination and viewpoint changes at the same time. We showed that disentangling the homography estimation from representation learning provides better estimates. We also proposed to use an additional photometric distortion step in the synthetic COCO dataset generation (PDS-COCO) and encourage future works to use it as a new evaluation benchmark of robust homography estimation. Then, we presented a study of modern homography estimation baselines along with our bidirectional implicit Homography Estimation (\textit{biHomE}) loss on both S-COCO and PDS-COCO. \textit{biHomE} loss achieves a new state-of-the-art performance for unsupervised homography estimation, which is also comparable or better compared to supervised approaches. Furthermore, we investigated the influence of image alignment using homography on change captioning quality. We showed that aligning the images using our \textit{biHomE} loss is not only beneficial for the overall captioning quality of DUDA architecture on the CLEVR\_Change dataset but also for performance degradation from smaller to bigger viewpoint shifts.

{\small
\bibliographystyle{ieee_fullname.bst}
\bibliography{egbib.bib}
}

\begin{appendices}


\begin{figure}[b]
    \centering
    \hspace{5pt}
    \includegraphics[width=0.38\columnwidth]{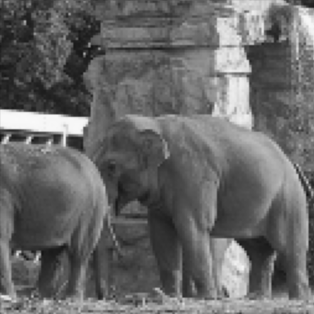}
    \includegraphics[width=0.38\columnwidth]{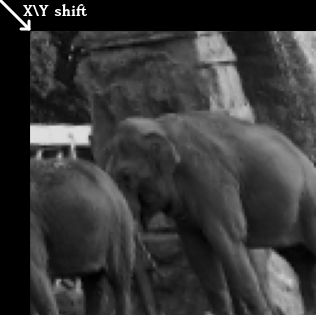}
    \includegraphics[width=\columnwidth]{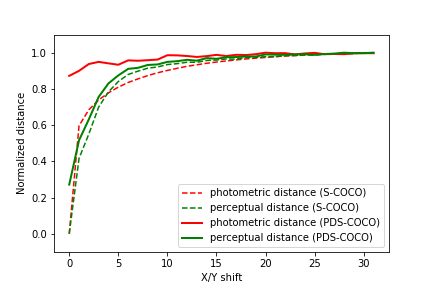}
    \caption{Normalized distance as a function of image shift. The Figure is prepared for one hundred random images from S-COCO (dashed) PDS-COCO dataset (solid), where the target image was shifted by a given number of pixels in both X and Y axes. We used pretrained ResNet34 up to the first residual block as perceptual \textit{Loss Network} and MSE as photometric distance. For S-COCO distance statistics is similar for both distances, but for PDS-COCO perceptual distance is smoother.}
    \label{fig::distance-by-viewpoint-shift}
\end{figure}


\begin{table*}[t]
\centering
\begin{tabular}{@{}lcccccc@{}}
\toprule
                        & RE                                                & LT                                                & LL                                                & SF                                                & LF                                                & Avg                                               \\ \midrule
Zhang (reported)        & 1.81                                              & 1.90                                              & 1.94                                              & 1.75                                              & 1.72                                              & 1.82                                              \\
Zhang (reproduced)      & $1.813$ \scriptsize{$\pm 0.013$} & $2.139$ \scriptsize{$\pm 0.013$} & $1.906$ \scriptsize{$\pm 0.013$} & $1.837$ \scriptsize{$\pm 0.010$} & $1.894$ \scriptsize{$\pm 0.006$} & $1.918$ \scriptsize{$\pm 0.006$} \\
\textit{Zhang + biHomE} & $1.822$ \scriptsize{$\pm 0.006$} & $2.178$ \scriptsize{$\pm 0.031$} & $1.924$ \scriptsize{$\pm 0.011$} & $1.842$ \scriptsize{$\pm 0.006$} & $1.994$ \scriptsize{$\pm 0.009$} & $1.941$ \scriptsize{$\pm 0.008$} \\ \bottomrule
\end{tabular}
\caption{The performance of the original Zhang \etal \cite{zhang2019content} method (reported in the paper and reproduced by us) and trained with our \textit{biHomE} loss on their dataset \cite{zhang2019content}. The performance of the original Zhang \etal \cite{zhang2019content} method is only slightly better than using our biHomE loss. We hypothesize this is because this dataset consists mostly of image pairs with small viewpoint and illumination changes.}
\label{tab::real-dataset}
\end{table*}


\section{Photometric or Perceptual}

In this Section, we want to additionally explore the possible reason behind the effectiveness of perceptual loss functions. Photometric loss is known to be sensitive to illumination conditions, while high-level features extracted from pretrained networks care more about perceptual similarity \cite{johnson2016perceptual, wang2018esrgan}. To better understand the behavior of both losses we prepare a simple experiment, where the target image is shifted by a given number of pixels in the X and Y axis w.r.t the source image. Then we report $L_1$ distance in pixel space and $L_1$ distance in feature space produced by the \textit{Loss Network} $g$ between both images. To bring both distances in the same range we normalize them by the maximum observed distance. The average of one hundred images for both S-COCO and PDS-COCO is presented in Figure \ref{fig::distance-by-viewpoint-shift}.

The distance curve of photometric loss and perceptual loss on S-COCO is similar, so we expect the comparable performance of both loss functions. However, when photometric distortion is introduced, the perceptual loss function is smoother and will likely produce better results. Indeed, both conclusions are supported by the illumination robustness experiments shown in Section \ref{sec::illumination_robustness} of the main paper.


\section{biHomE Performance on Real-World Dataset}

The dataset proposed by Zhang \etal \cite{zhang2019content} is composed of 80$k$ image pairs extracted from short video clips containing small camera movements and dynamic objects on the scene. The image pairs are divided into 5 categories: regular (RE), low-texture (LT), low-light (LL), small-foregrounds (SF), and large-foreground (LF) scenes. We reproduced the original Zhang \etal \cite{zhang2019content} method and using the same learning setting we also learned our \textit{biHomE} loss.

As one can observe in Table \ref{tab::real-dataset} the performance of the original Zhang \etal \cite{zhang2019content} method is only slightly better than using our \textit{biHomE} loss. A similar effect could be observed also in Section \ref{sec::illumination_robustness}, where for small viewpoint change $\rho$ and small photometric distortion $\delta$ original Zhang method is also better than with our \textit{biHomE} loss. 

\end{appendices}
\end{document}

%% file: figure_tikz.tex

\pgfplotsset{
    bar group size/.style 2 args={
        /pgf/bar shift={%
                -0.5*(#2*\pgfplotbarwidth + (#2-1)*\pgfkeysvalueof{/pgfplots/bar group skip})  + 
                (.5+#1)*\pgfplotbarwidth + #1*\pgfkeysvalueof{/pgfplots/bar group skip}},%
    },
    bar group skip/.initial=2pt,
    plot 0/.style={blue,fill=blue!40!white,mark=none},%
    plot 1/.style={green!60!black,fill=green!40!white,mark=none},%
    plot 2/.style={brown!60!black,fill=brown!40!white,mark=none},%
    plot 3/.style={red,fill=red!40!white,mark=none},%
}

\pgfdeclarepatternformonly{mystars}
  {\pgfpointorigin}
  {\pgfqpoint{1.5mm}{1.5mm}}
  {\pgfqpoint{1.5mm}{1.5mm}}
  {
   \pgftransformshift{\pgfqpoint{1mm}{0.5mm}}
   \pgfpathmoveto{\pgfqpointpolar{18}{0.5mm}}
   \pgfpathlineto{\pgfqpointpolar{162}{0.5mm}}
   \pgfpathlineto{\pgfqpointpolar{306}{0.5mm}}
   \pgfpathlineto{\pgfqpointpolar{90}{0.5mm}}
   \pgfpathlineto{\pgfqpointpolar{234}{0.5mm}}
   \pgfpathclose%
   \pgfusepath{fill}
  }

\begin{figure}[t]
\centering


\begin{tikzpicture}

  \begin{groupplot}[group style={group size=2 by 1, horizontal sep=2.6cm},height=2.3cm,width=0.5\columnwidth]
  
  
    \nextgroupplot[
    xbar,
    y axis line style = { opacity = 0 },
    axis x line       = none,
    tickwidth         = 0pt,
    symbolic y coords = {DeTone, Zeng},
    nodes near coords,
    every node near coord/.append style={xshift=-25pt,
        /pgf/number format/fixed zerofill,
        /pgf/number format/precision=2},
    xscale=-1, yscale=1,
    yticklabels={,,}, 
    clip=false,
    xmin=0,
    xmax=7,
    legend style={at={(3pt, -40pt)},anchor=south,legend columns=2},
    legend image code/.code={\draw[#1, draw=none] (0cm,-0.1cm) rectangle (0.3cm,0.1cm);},  
    ]
    \addplot[plot 0, bar group size={0}{1}, postaction={pattern=dots},
    error bars/.cd,
    x dir=both,
    x explicit,
    error bar style={plot 0,line width=1pt},
    error mark options={
      rotate=90,
      blue,
      mark size=2pt,
      line width=1pt
    }]
    coordinates {
    (1.96,DeTone) +- (0.12,0.12)
    (1.73,Zeng) +- (0.1,0.1)
    };
    \node (title) at (18pt, 32pt) {S-COCO};
    
    
    \nextgroupplot[
    xbar,
    y axis line style = { opacity = 0 },
    axis x line       = none,
    tickwidth         = 0pt,
    symbolic y coords = {DeTone,Zeng},
    yticklabel style={text width=2.85cm, align=center, xshift=4pt},
    ytick=\empty,
    extra y ticks = {DeTone,Zeng},
    nodes near coords,
    every node near coord/.append style={xshift=0pt,
    /pgf/number format/fixed zerofill,
    /pgf/number format/precision=2},
    clip=false,
    xmin=0,
    xmax=7,
    ]
    
    \addplot[plot 0, bar group size={0}{1}, postaction={pattern=dots},
    error bars/.cd,
    x dir=both,
    x explicit,
    error bar style={plot 0,line width=1pt},
    error mark options={
      rotate=90,
      blue,
      mark size=2pt,
      line width=1pt
    }]
    coordinates {
    (2.50,DeTone) +- (0.20,0.20)
    (2.27,Zeng) +- (0.07,0.07)
    };
    \node (title) at (24pt, 32pt) {PDS-COCO};
  \end{groupplot}

  \draw[thick, dashed] (0.9,-0.25) -- (6.875,-0.25);
  
\end{tikzpicture}


\begin{tikzpicture}

  \begin{groupplot}[group style={group size=2 by 1, horizontal sep=2.6cm},height=2.3cm,width=0.5\columnwidth]
  
  
    \nextgroupplot[
    xbar,
    y axis line style = { opacity = 0 },
    axis x line       = none,
    tickwidth         = 0pt,
    symbolic y coords = {Zhang, Nguyen},
    nodes near coords,
    every node near coord/.append style={xshift=-25pt,
        /pgf/number format/fixed zerofill,
        /pgf/number format/precision=2},
    xscale=-1, yscale=1,
    yticklabels={,,}, 
    clip=false,
    xmin=0,
    xmax=7,
    ]

    \addplot[plot 0, bar group size={0}{1}, postaction={pattern=dots},
        error bars/.cd,
        x dir=both,
        x explicit,
        error bar style={brown!60!black,line width=1pt},
        error mark options={
      rotate=90,
      brown!60!black,
      mark size=2pt,
      line width=1pt}] coordinates { (2.07,Nguyen)+-(0.1,0.1) (2.08,Zhang)+-(0.11,0.11) };
    

    
    \nextgroupplot[
    xbar,
    y axis line style = { opacity = 0 },
    axis x line       = none,
    tickwidth         = 0pt,
    symbolic y coords = {Zhang, Nguyen},
    yticklabel style={text width=2.85cm, align=center, inner sep=0pt, xshift=4pt},
    ytick=\empty,
    extra y ticks = {Zhang, Nguyen},
    nodes near coords,
    every node near coord/.append style={xshift=0pt,
    /pgf/number format/fixed zerofill,
    /pgf/number format/precision=2},
    clip=false,
    xmin=0,
    xmax=7,
    ]
    
    \addplot+[plot 3,bar group size={0}{1},nodes near coords={},postaction={pattern=mystars}] coordinates { (5.00,Nguyen) (5.00,Zhang) };
    
  \end{groupplot}
  
  \draw[thick, dashed] (0.9,-0.25) -- (6.875,-0.25);
  
\end{tikzpicture}


\begin{tikzpicture}

  \begin{groupplot}[group style={group size=2 by 1, horizontal sep=2.6cm},height=3.0cm,width=0.5\columnwidth]
  
  
    \nextgroupplot[
    xbar,
    y axis line style = { opacity = 0 },
    axis x line       = none,
    tickwidth         = 0pt,
    symbolic y coords = {Zhang+biHomE, DeTone+biHomE, Zeng+biHomE},
    nodes near coords,
    every node near coord/.append style={xshift=-25pt,
        /pgf/number format/fixed zerofill,
        /pgf/number format/precision=2},
    xscale=-1, yscale=1,
    yticklabels={,,}, 
    clip=false,
    xmin=0,
    xmax=7,
    ]

    \addplot[plot 1, bar group size={0}{1}, postaction={pattern=north east lines},
        error bars/.cd,
        x dir=both,
        x explicit,
        error bar style={green!60!black,line width=1pt},
        error mark options={
      rotate=90,
      green!60!black,
      mark size=2pt,
      line width=1pt}]
    coordinates {(1.82,DeTone+biHomE)+-(0.06,0.06) (1.79,Zeng+biHomE)+-(0.05,0.05) (1.87,Zhang+biHomE)+-(0.07,0.07)};

    
    \nextgroupplot[
    xbar,
    y axis line style = { opacity = 0 },
    axis x line       = none,
    tickwidth         = 0pt,
    symbolic y coords = {Zhang+biHomE, DeTone+biHomE, Zeng+biHomE},
    yticklabel style={text width=2.85cm, align=center, inner sep=0pt, xshift=4pt},
    ytick=\empty,
    extra y ticks = {Zhang+biHomE, DeTone+biHomE, Zeng+biHomE},
    nodes near coords,
    every node near coord/.append style={xshift=0pt,
    /pgf/number format/fixed zerofill,
    /pgf/number format/precision=2},
    clip=false,
    xmin=0,
    xmax=7,
    legend style={at={(-27pt, -42pt)},anchor=south,legend columns=2},
    legend image code/.code={\draw[#1, draw=none] (+0cm,-0.1cm) rectangle (0.3cm,0.1cm);},
    ]
    
    \addplot[plot 1, bar group size={0}{1}, postaction={pattern=north east lines},
        error bars/.cd,
        x dir=both,
        x explicit,
        error bar style={green!60!black,line width=1pt},
        error mark options={
      rotate=90,
      green!60!black,
      mark size=2pt,
      line width=1pt}]
    coordinates {(2.39,DeTone+biHomE)+-(0.20,0.20) (2.11,Zeng+biHomE)+-(0.06,0.06) (2.51,Zhang+biHomE)+-(0.1,0.1)};
    
  \end{groupplot}
  
\end{tikzpicture}


\begin{tikzpicture}
    \begin{axis}[%
    hide axis,
    xmin=0,
    xmax=1,
    ymin=0,
    ymax=1,
    height=1.7cm,
    width=5cm,
    legend style={at={(0.5,0.0)},anchor=north,legend columns=3},
    legend image code/.code={\draw[#1, draw=none] (+0cm,-0.1cm) rectangle (0.3cm,0.1cm);},
    ]
    \addlegendimage{plot 0,postaction={pattern=dots}}
    \addlegendentry{Reproduced~~~~~~~~};
    \addlegendimage{plot 3,postaction={pattern=mystars}}
    \addlegendentry{Not Converged~~~~~~~~};
    \addlegendimage{plot 1,postaction={pattern=north east lines}}
    \addlegendentry{Ours};
    
    \end{axis}
\end{tikzpicture}


\begin{tikzpicture}[remember picture,overlay,baseline]
  \node[draw,rotate=90] at (-3.4cm,2.55cm) {\textcolor{black}{~~~~unsupervised~~~~}};
  \node[draw,rotate=90] at  (-3.4cm,5cm) {supervised};
\end{tikzpicture}

\caption{Homography estimation comparison on synthetic COCO (S-COCO)~\cite{detone2016deep} on the left and a photometrically distorted version of it (PDS-COCO) on the right. On PSD-COCO, our bidirectional implicit Homography Estimation (\textit{biHomE}) loss is the only unsupervised method able to converge, while still being on par with the performance of supervised approaches. The performance of methods (Nguyen~\cite{nguyen2018unsupervised}, DeTone~\cite{detone2016deep}, Zeng~\cite{zeng2018rethinking}, and Zhang~\cite{zhang2019content}) is reported in Mean Absolute Corner Error (MACE).
}

\label{fig::exp-1}
\end{figure}